\documentclass[letterpaper]{article} 
\usepackage{aaai2026}  
\usepackage{times}  
\usepackage{helvet}  
\usepackage{courier}  
\usepackage[hyphens]{url}  
\usepackage{graphicx} 
\usepackage{natbib}  
\usepackage{caption} 
\usepackage{amssymb}
\usepackage{amsmath}
\usepackage{algorithm}
\usepackage{algpseudocode}
\algrenewcommand\algorithmicrequire{\textbf{Input:}}
\algrenewcommand\algorithmicensure{\textbf{Output:}}
\usepackage{subcaption}
\usepackage{xcolor}

\newcommand{\edit}[1]{\textcolor{black}{#1}}

\usepackage{newfloat}
\usepackage{listings}
\DeclareCaptionStyle{ruled}{labelfont=normalfont,labelsep=colon,strut=off} 
\floatstyle{ruled}
\newfloat{listing}{tb}{lst}{}
\floatname{listing}{Listing}
\title{Optimizing Product Provenance Verification using Data Valuation Methods}

\author {
    Raquib Bin Yousuf\textsuperscript{\rm 1},
    Hoang Anh Just\textsuperscript{\rm 1},
    Shengzhe Xu\textsuperscript{\rm 1},
    Brian Mayer\textsuperscript{\rm 1},\\
    Victor Deklerck\textsuperscript{\rm 2},
    Jakub Truszkowski\textsuperscript{\rm 3,\rm 4},
    John C. Simeone\textsuperscript{\rm 5},
    Jade Saunders\textsuperscript{\rm 4},\\
    Chang-Tien Lu\textsuperscript{\rm 1},
    Ruoxi Jia\textsuperscript{\rm 1},
    Naren Ramakrishnan\textsuperscript{\rm 1}
}
\affiliations {
    \textsuperscript{\rm 1}Virginia Tech, VA, USA. 
    \textsuperscript{\rm 2}Meise Botanic Garden, Meise, Belgium\\
    \textsuperscript{\rm 3}Chalmers University of Technology, Gothenburg, Sweden\\
    \textsuperscript{\rm 4}World Forest ID, Washington, DC, USA\\
    \textsuperscript{\rm 5}Simeone Consulting, LLC, Littleton, NH, USA\\
    raquib@vt.edu, naren@cs.vt.edu
}

\begin{document}
\maketitle
\begin{abstract}
Determining and verifying product provenance remains a critical challenge in global supply chains, particularly as geopolitical conflicts and shifting borders create new incentives for misrepresentation of commodities, such as hiding the origin of illegally harvested timber or stolen agricultural products. Stable Isotope Ratio Analysis (SIRA), combined with Gaussian process regression-based isoscapes, has emerged as a powerful tool for geographic origin verification. While these models are now actively deployed in operational settings supporting regulators, certification bodies, and companies, they remain constrained by data scarcity and suboptimal dataset selection.
In this work, we introduce a novel deployed data valuation framework designed to enhance the selection and utilization of training data for machine learning models applied in SIRA. By quantifying the marginal utility of individual samples using Shapley values, our method guides strategic, cost-effective, and robust sampling campaigns within active monitoring programs. By prioritizing high-informative samples, our approach improves model robustness and predictive accuracy across diverse datasets and geographies. 
Our framework has been implemented and validated in a live provenance verification system currently used by enforcement agencies, demonstrating tangible, real-world impact. Through
extensive experiments and deployment in a live provenance verification system, we show that this system significantly enhances provenance verification, mitigates fraudulent trade practices, and strengthens regulatory enforcement of global supply chains.
\end{abstract}
\section{Introduction}
Global natural resource supply chains are opaque,  especially since natural resources are often transformed from raw materials (e.g., timber) into finished consumer-facing products (e.g., furniture). These complex supply chains often involve multiple countries, with intermediate outputs being traded internationally and being used as inputs into further manufacturing processes. In addition to business-commerce decisions driving supply chain sourcing, the economics of natural resource trade are often closely linked with geopolitics. 
\edit{Verifying the true origin of products remains difficult and high-stakes,}
as geopolitical incentives and ``don't ask, don't tell'' sourcing cultures encourages misrepresentation of commodities, particularly for illegally harvested timber or stolen agricultural products.

\paragraph{The Scale of the Problem: Illegal Timber as a Case Study.}
Illegal logging, in particular, is the most profitable natural resource crime, valued at US\$52 billion to US\$157 billion per year~\cite{transnational}. Illegally obtained timber accounts for 10--30\% of the total global trade in timber products, and in regions such as Southeast Asia, Central Africa, and South America, it is estimated that 50--90\% of timber is harvested illegally~\cite{transnational}. \edit{These figures underscore the urgent need for deployed, verifiable, and scientifically defensible provenance systems.}

\paragraph{Forensic and Analytical Tools for Provenance Verification.}
Product identification and provenance verification of traded natural resources have emerged as promising research areas, with various combinations of methods used depending on the resource sector and the granularity required for species identification and origin determination; for example, species and geographic harvest provenance for wood and forest products often requires multiple forensic tools and testing methods~\cite{wood_forensic, schmitz:hal-02936035, dormontt2015forensic}. For geographic origin verification in particular, Stable Isotope Ratio Analysis (SIRA), combined with Gaussian process regression-based isoscapes, has proven highly effective~\cite{truszkowski2025, mortier2024framework}. SIRA leverages the natural variation in stable isotope ratios—measured via mass spectrometry—to determine the enrichment of non-radioactive elemental isotopes in a sample~\cite{barrie1996automated}, with enrichment patterns driven by environmental, atmospheric, soil, metabolic, and species-specific factors \cite{siegwolf_stable_2022, wang2021possible, vystavna2021temperature}. It has been successfully used to trace the origin of timber, seafood, agricultural products, and fiber such as cotton \cite{truszkowski2025, mortier2024framework, watkinson2022case, cusa_future_2022, wang2020tracing, meier-augenstein_discrimination_2014}.
However, \edit{operational deployment of these models faces challenges of data scarcity, suboptimal reference selection, and high sampling costs,} motivating the need for systematic data valuation and prioritization. This in turn highlights the
importance of optimizing and `valuing'
reference sample collection efforts~\cite{gasson2021worldforestid}.

\paragraph{Operationalizing Provenance Verification: The WFID Platform.}
Efforts such as World Forest ID (WFID) (worldforestid.org) have operationalized geographic origin models to address this problem, particularly in the timber trade. WFID identifies high-risk species and geographies \cite{birch_wf_2}, collects chain-of-custody reference samples, and uses laboratory analysis \cite{fera_science_3} to generate the chemical data needed for stable isotope modeling.
\edit{These models are not only research prototypes—they are currently deployed and used in live compliance systems.}
The World Forest ID Evaluation Platform \cite{wf_evaluation_4} allows regulators, certification bodies, and companies to evaluate sourcing claims in real time.

Here, we introduce our \edit{deployed} data valuation framework that enhances the selection and utilization of training data for machine learning models applied to SIRA. Field sample collections, central to any scientific traceability method, are logistically difficult and expensive due to remote locations, specialized equipment, and labor-intensive workflows.
\edit{Our deployed system operationalizes Shapley-based data valuation} to quantify the marginal utility of individual samples, enabling strategic, cost-effective, and robust sampling campaigns that \edit{directly improve model accuracy in active enforcement workflows.}

\paragraph{Contributions:}
\begin{enumerate}
    \item \edit{Application of machine learning data valuation to an operational provenance verification context.} 
By prioritizing highly informative samples, our approach improves model robustness and predictive accuracy across diverse datasets and geographies. 
\item \edit{Deployment in live enforcement systems.}
We have deployed our optimized sampling approach in a provenance verification system used by European enforcement agencies to curb the trade of sanctioned Russian timber by proving that alternate claimed origins are non-viable.
See coverage of our work in the \textit{New York Times}~\cite{nytimes_nazaryan_2024}).
Due to confidentiality reasons, we use a global dataset of Oak ({\it Quercus spp.}) reference samples to illustrate our methodology.
\item  \edit{Extensive empirical validation and field integration.} 
We validate our framework with global \edit{Oak (Quercus spp.) reference datasets}, demonstrating its potential to enhance data valuation, optimize model configuration, and \edit{strengthen the regulatory enforcement of global supply chains.}
\end{enumerate}

\section{Related Work}
SIRA has been widely employed as a geographical discriminator for various plant and animal-based products in global supply chains, such as garlic \cite{pianezze_geographical_2019}, Chinese tea \cite{liu_c_2020}, olive oil \cite{bontempo_characterisation_2019}, cheese \cite{camin_application_2004}, and timber \cite{mortier2024framework, truszkowski2025}. 
By bringing data valuation methods to bear upon
SIRA pipelines we aim to improve
the verification of product provenance.

Prior work in data valuation is typically
seen in the context of explainable machine learning and enhancing model performance~\cite{wu2024data, covert2024stochastic}. Existing methods primarily rely on leave-one-out retraining and influence functions \cite{koh2017understanding}, Shapley values \cite{jia2019towards, ghorbani2019data, shapley_jia_wang2023note}, Least Cores \cite{yan2021if}, the Banzhaf value \cite{banzhaf_wang2022data}, Beta Shapley \cite{kwon2021beta}, and reinforcement learning \cite{yoon2020data}. Furthermore, data valuation has been applied across various domains to enhance model development and interpretability, including health data \cite{health_pandl2021trustworthy}, medical imaging \cite{medical_image_tang2021data}, and the Internet of Things \cite{iot_shi2024data}. This paper
is the first to formally apply data
valuation techniques to SIRA.

\section{Methods}
\label{sec:methods}
Let $\mathcal{X}$ be a set of locations where
data is collected; $x\in \mathcal{X}$ typically is specified by a longitude and latitude. Let $\mathcal{Y}$ be the set of measurements made
over $\mathcal{X}$, here 
denoting stable isotope ratio values (e.g. $\delta^{13}C$, $\delta^2H$, $\delta^{15} N$, $\delta^{18}O$, $\delta^{34}S$), or trace element values (e.g. Si, Cu, S, Ba, Rb).
We denote $f$ and $g$ as functions of interest (defined below). For evaluation purposes, we split our data into training and test datasets, where $D=\{z_i\}_{i=1}^N$ represents the training dataset with $N$ data points and $z_i = (x_i, y_i) \in \mathcal{X} \times \mathcal{Y}$. Similarly, $T=\{z_i\}_{i=1}^M$ denotes the test dataset with $M$ data points. We let $v(h, A, B)$ denote the performance of the model $h$ trained on a dataset $A$ and evaluated on the dataset $B$, where $v$ would return a numerical value, $v(h, A, B) \in \mathbb{R}$. In cases when the function and the test dataset are known, we drop the dependence in the notation to simply say $v(A)$.

\subsection{Forward and Backward Models}
\paragraph{\textbf{Forward Models.}} For the forward model, the task is to predict the stable isotope values for a given location, i.e., $f : \mathcal{X} \rightarrow \mathcal{Y}$. The motivation is to verify whether the characteristics of the location (denoting specified harvest origin) would align with general isotopic values associated with the specified location. Such models can be based on approahes like decision trees, random forests, or XGBoost. Recent works have proposed using Gaussian process regression models with high performance~\cite{truszkowski2025, mortier2024framework}. 

\paragraph{\textbf{Backward Models.}} In the backward case, we aim to identify the location given
measured stable isotope values. We model this relation as $g : \mathcal{Y} \rightarrow \mathcal{X}$. The fitted model $g$ would help identify whether the declared harvest location of a species sample aligns with the predicted location from $g$. Similarly, here, one can use a range of machine learning models to fit this function. For instance, \citet{mortier2024framework} reversed a fitted Gaussian process regression model using the Bayes' rule to predict locations from measured isotope ratios.

\paragraph{\textbf{Atmospheric Variables.}} In addition, to support either forward or backward models, we often have available a range of atmospheric variables associated with locations. Such variables can be used as either additional inputs to a forward model or auxiliary information in a backward model.

\paragraph{\textbf{Gaussian Process Regression Models.}} Gaussian process regression (GPR) models, as explored by \cite{truszkowski2025, mortier2024framework}, offer a powerful approach for both forward and backward modeling settings.
For the forward model $f: \mathcal{X} \rightarrow \mathcal{Y}$, GPR can be used to predict isotope ratios $y \in \mathcal{Y}$ at a given location $x \in \mathcal{X}$. We can construct ``isoscapes'' by fitting independent GP regression models to each feature in $\mathcal{Y}$.  Considering our training dataset $D=\{z_i\}_{i=1}^N = \{(x_i, y_i)\}_{i=1}^N$, for each feature $y_j \in \mathcal{Y}$, a GP model is trained to predict isotope values at a new location $x^* \in \mathcal{X}$. The predicted distribution for the feature $y_j$ at location $x^*$ is Gaussian, with a mean and variance given by:
\begin{align*}
E[y_j|x^*, \mathbf{X}] = & \mu_j + \mathbf{k}^{(j)T}(\mathbf{K}^{(j)} + \sigma_j^2\mathbf{I})^{-1}(\mathbf{y}_j - \mu_j)\\
V(y_j|x^*, \mathbf{X}) =\; & k^{(j)}(\mathbf{x^*}, \mathbf{x^*}) + \sigma_j^2 \\
& - \mathbf{k}^{(j)T}(\mathbf{K}^{(j)} + \sigma_j^2 \mathbf{I})^{-1} \mathbf{k}^{(j)}
\end{align*}

Here, $\mathbf{X} = \{x_1, \dots, x_N\}$ represents the training locations and $\mathbf{y}_j = [y_{1j}, \dots, y_{Nj}]^T$ are the values of the $j$-th feature in the training set. $\mu_j$ is the baseline mean for feature $y_j$, $\mathbf{K}^{(j)}$ is the covariance matrix evaluated at all pairs of training locations, $\mathbf{k}^{(j)}$ is the covariance vector between the test location $x^*$ and the training locations, $k^{(j)}(x^*, x^*)$ is the covariance of $x^*$ with itself, and $\sigma_j^2$ is the noise variance for feature $y_j$.  For the backward model $g: \mathcal{Y} \rightarrow \mathcal{X}$, we leverage Bayesian inference to reverse the prediction. Given a set of features $y^* \in \mathcal{Y}$ from a location of unknown origin, the posterior probability of its origin being location $x^* \in \mathcal{X}$ is calculated using Bayes' theorem:
$$ p(x^*|y^*, D) = \frac{p(y^*|x^*, D)p(x^*)}{\int_{x \in \mathcal{X}} p(y^*|x, D)p(x) dx}$$
The likelihood $p(y^*|x^*, D)$ is derived from the forward GP model, assuming independence of features and using the predicted Gaussian distributions:
\begin{align*}
p(y^* \mid x^*, D) 
&= \prod_{j \in \mathcal{Y}} 
    \frac{1}{\sqrt{2\pi V(y_j \mid x^*, D)}} \nonumber \\
&\quad \times 
    \exp\left(
        -\frac{
            (y_j^* - E[y_j \mid x^*, D])^2
        }{
            2V(y_j \mid x^*, D)
        }
    \right)
\end{align*}
The prior $p(x^*)$ can incorporate prior knowledge about the distribution of tree harvest locations. This Bayesian approach provides a posterior probability map over $\mathcal{X}$, indicating the most likely origin locations for an observation with features $y^*$.  The performance of both forward and backward GPR models can be assessed using the metric $v(h, D, T)$, where $h$ is the GPR model ($f$ or $g$).

\paragraph{\textbf{Performance Metrics.}} 
The primary metric we will employ is:

\[
\text{RMSE} = \sqrt{\int_{\mathbf{x} \in A} \big( d(\mathbf{x}_t, \mathbf{x}) \big)^2 \, p(\mathbf{x}|\mathbf{y}^*, D) \, d\mathbf{x}},
\]
\noindent
where \( d(\mathbf{x}_t, \mathbf{x}) \) is the great circle distance. 
Comparing RMSE across different GP models helps identify which model minimizes large prediction errors and provides overall reliable estimates.
\subsection{Data Valuation}
The  Shapley value, introduced by \cite{shapley1953value}, offers a principled approach to quantify data value, identifying both highly informative and potentially detrimental data points. The Shapley value $\phi_i$ for a data point $i$ is computed as the weighted average of its marginal contribution to model performance across all possible subsets of the training data: 
$$\phi_i = \sum_{S \subseteq D \setminus \{z_i\}} \frac{|S|!(|D| - |S| - 1)!}{|D|!} [v(S \cup \{z_i\}) - v(S)].$$
Here, $D$ is the full training set, $S$ is a subset excluding $i$, and $v(S)$ is the model performance (e.g., negative mean absolute error) when trained on subset $S$. High positive Shapley values indicate highly valuable data points that significantly improve performance, while low or negative values suggest redundancy or detrimental effects, possibly due to outliers, measurement errors, or model misspecification. The Shapley value is not an arbitrary metric; it is uniquely characterized as that satisfying a set of desirable axioms, ensuring fairness and consistency in data valuation:

\begin{enumerate}
\item   \textbf{Efficiency:} The sum of the Shapley values for all data points equals the difference in performance between the model trained on the full dataset and the model trained on an empty dataset: $\sum_{i \in D} \phi_i = v(D) - v(\emptyset)$. This means the total value is fully distributed among the data points.
\item   \textbf{Symmetry (or Null Player):} If a data point $i$ has zero marginal contribution to every possible subset (i.e., $v(S \cup \{i\}) = v(S)$ for all $S$), then its Shapley value is zero: $\phi_i = 0$. Useless data points receive zero value.
\item   \textbf{Linearity:} If the performance metric $v$ is a linear combination of two other performance metrics, $v = a \cdot v_1 + b \cdot v_2$, then the Shapley values for $v$ are the same linear combination of the Shapley values for $v_1$ and $v_2$. This ensures consistency across different performance measures.
\item   \textbf{Dummy:} if two data points $i$ and $j$ always have the same marginal contribution to every subset of $D$ then their shapely value must be equal. $\phi_i = \phi_j$.
\end{enumerate}
These axioms provide a strong theoretical justification for using the Shapley value. Furthermore, the Shapley value  can be equivalently expressed as a sum over permutations of the dataset \citep{shapley1953value}:
$$
\phi_i = \sum_{\pi \sim \Pi(D)} \left[ v(P^{\pi}_i \cup z_i) - v(P^{\pi}_i ) \right],
$$
where $\Pi(D)$ is the set of all permutations of data points in $D$, $\pi$ is a permutation sampled uniformly at random from $\Pi(D)$, and $P^{\pi}_i$ is the set of data points preceding instance $z_i$ in permutation $\pi$.
\edit{This permutation-based form is equivalent to the subset definition and underlies the Monte Carlo approximation \cite{ghorbani2019data, kwon2021beta}.
In practice, we use the permutation form primarily for intuition, while our actual computations employ the subset-sampling TMC-Shapley method.}

\textbf{Truncated Monte Carlo Shapley Value.} Because the exact computation of Shapley values is computationally prohibitive,
\cite{ghorbani2019data} propose
their approximation
by randomly sampling a limited number of subsets instead of exhaustively considering all possibilities (see Algorithm 1 in the extended version). The key idea is that each random permutation provides an unbiased estimate of the marginal contribution of each data point,
convergence achievable in practice in $O(n)$ permutations (typically around $3n$) for an $n$-point dataset. Generally, Monte Carlo estimators exhibit variance that decreases proportional to $1/\sqrt{m}$ as the number of samples $m$ increases. However, in SIRA settings, each marginal evaluation entails retraining a Gaussian process with $\mathcal{O}(N^3)$ time complexity, creating a pronounced trade-off between computational feasibility and valuation precision. This tension motivates careful calibration of the permutation budget to balance estimator fidelity against real-world processing constraints.

\textbf{Iterative Data Selection with Shapley Values.}
To strategically select a subset $D' \subseteq D$ from the original training data $D$ that maximizes model accuracy for both forward and backward prediction tasks, we propose to leverage Shapley values computed once on the full dataset to identify data point importance. We hypothesize that by using these initial global valuations, we can efficiently identify and sequentially remove less valuable samples. Our data selection methodology consists of three distinct steps. Initially, we compute data values for all points in $D$ using the entire dataset. Subsequently, we sort the data points and remove the least valuable one according to these pre-calculated values. Finally, we evaluate the model's performance using the test dataset. This sequential removal process continues as long as model performance improves, relying on the initial single valuation (see Algorithm 2 in the extended version). 

\textbf{Beta Shapley.}
Building upon the foundational principles of the Shapley value, Beta Shapley offers a flexible generalization for data valuation by recognizing that the standard Shapley value’s uniform weighting of a data point’s marginal contribution across all subset sizes is not always optimal. In many data valuation tasks, the primary objective is to rank data points rather than precisely distribute the total model performance gain, which makes the strict efficiency axiom less critical. Beta Shapley therefore relaxes this axiom and introduces a weighted-average framework in which weights are governed by a Beta distribution. This enables a more fine-grained valuation by assigning different levels of importance to a data point’s marginal contribution depending on the cardinality of the subset it is added to.
For 
instance, by selecting appropriate parameters for the Beta
distribution, one can prioritize the contributions made to
smaller subsets (see Algorithm 3 in the extended version).
\edit{We include Beta Shapley primarily to provide conceptual background within the broader family of Shapley-based data valuation methods. Our focus remains on TMC-Shapley, given its superior scalability and lower computational variance in practice.}

\section{Experiments}
We utilized data from two datasets of the genus {\it Quercus}, collected from various regions worldwide. 
\edit{Two broad datasets were combined in this study
with the first dataset comprising
tree samples distributed globally ($N=491$), while the second dataset was
focuses specifically on European countries ($N=287$)}.
Stable isotope ratio measurements were performed following the protocols outlined in \citep{watkinson2020development, boner2007stable}. 
Our experiments are aimed at answering the below questions:
\begin{enumerate}
    \item RQ1: What is the role of data Shapley values in the domain of SIRA? (Section \ref{ssec:0_dist})
    \item RQ2: How does model architecture influence data Shapley values and the performance of the proposed data valuation framework? (Section \ref{ssec:1_gp_rf})
    \item RQ3: How does Shapley-based data selection compare against naive or exhaustive baselines? (Section \ref{ssec:RQ3_baseline})
    \item RQ4: Can the proposed data valuation framework enhance outcomes in cases involving data imputation for missing or noisy data? (Section \ref{ssec:2_imputation})
    \item RQ5: Can data selection methods based on data valuation improve the performance of both directions in SIRA? (Section \ref{ssec:3_direction})
    \item RQ6: Given a specific model and data valuation framework, what level of granularity is optimal for effective data selection? (Section \ref{ssec:4_clustering})
    \item RQ7: Do different genera and species within the dataset exhibit varying data Shapley values? Can we identify the most and least important genera or species within the dataset? (Section \ref{ssec:5_species})
\end{enumerate}
\begin{figure}[hbtp]
    \centering
    \includegraphics[width=0.75\linewidth]{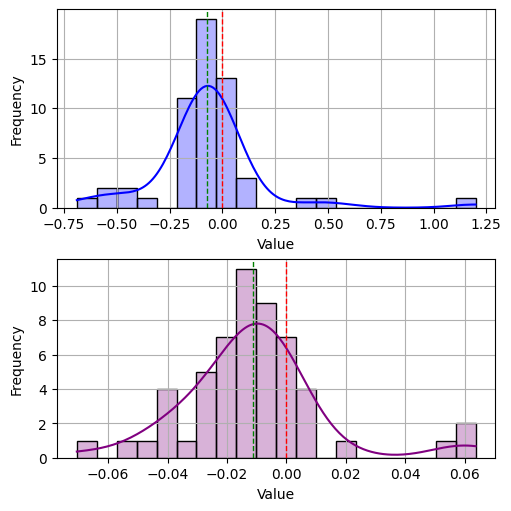}
    \caption{\edit{Distribution of data Shapley values, green and red lines represent the mean and the zero, respectively (top: backward model, bottom: forward model); see Section \ref{ssec:0_dist}.}}
    \label{fig:dist}
\end{figure}
\begin{figure*}[t]
\centering
    \begin{subfigure}[t]{0.30\linewidth}
        \centering
        \includegraphics[width=\linewidth]{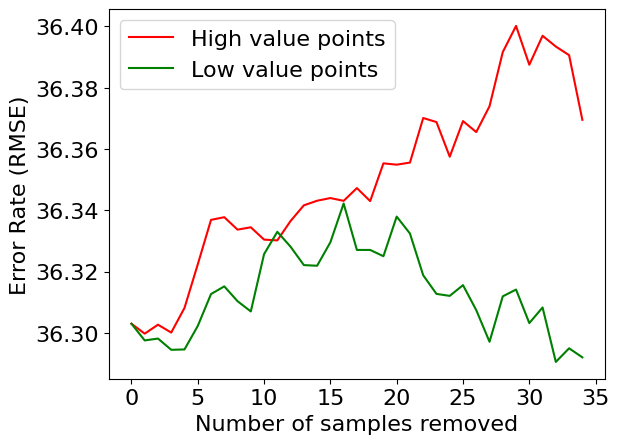}
        \caption{Error plot for Gaussian process regression}
        \label{fig:gp_error}
    \end{subfigure}
    \hfill
    \begin{subfigure}[t]{0.30\linewidth}
        \centering
        \includegraphics[width=\linewidth]{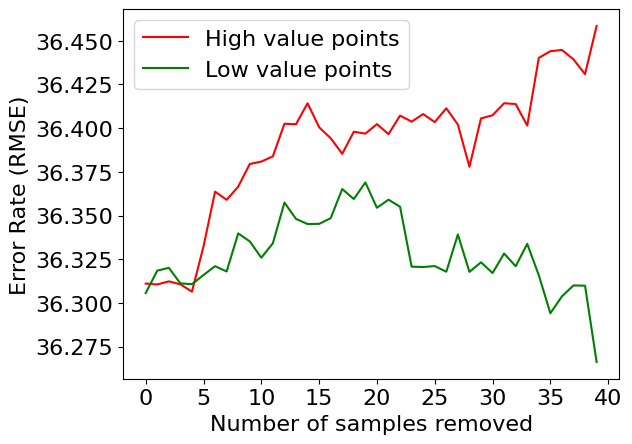}
        \caption{Error plot for random forest}
        \label{fig:rf_error_gp}
    \end{subfigure}
    \hfill
    \begin{subfigure}[t]{0.30\linewidth}
        \centering
        \raisebox{0.20\height}{\includegraphics[width=\linewidth]{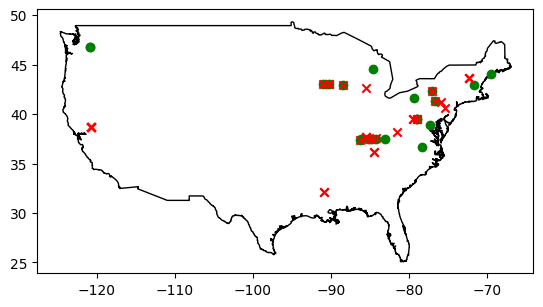}}
        \caption{Data selection on a map with Gaussian process regression (perturbed for privacy)}
        \label{fig:map_gp}
    \end{subfigure}
    \begin{subfigure}[b]{0.30\linewidth}
        \centering
        \includegraphics[width=\linewidth]{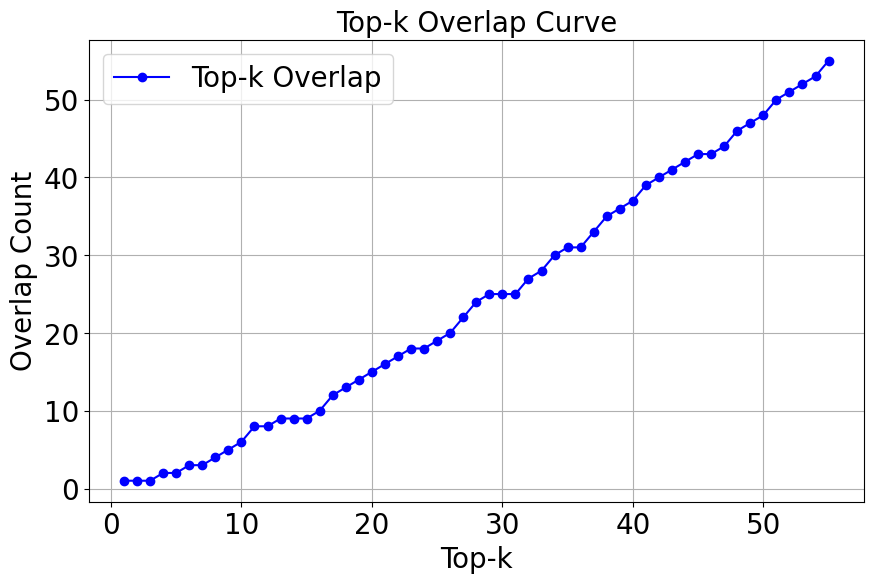}
        \caption{Data Shapley rank overlap with Gaussian process regression and random forest}
    \end{subfigure}
    \hfill
    \begin{subfigure}[b]{0.30\linewidth}
        \centering
        \includegraphics[width=\linewidth]{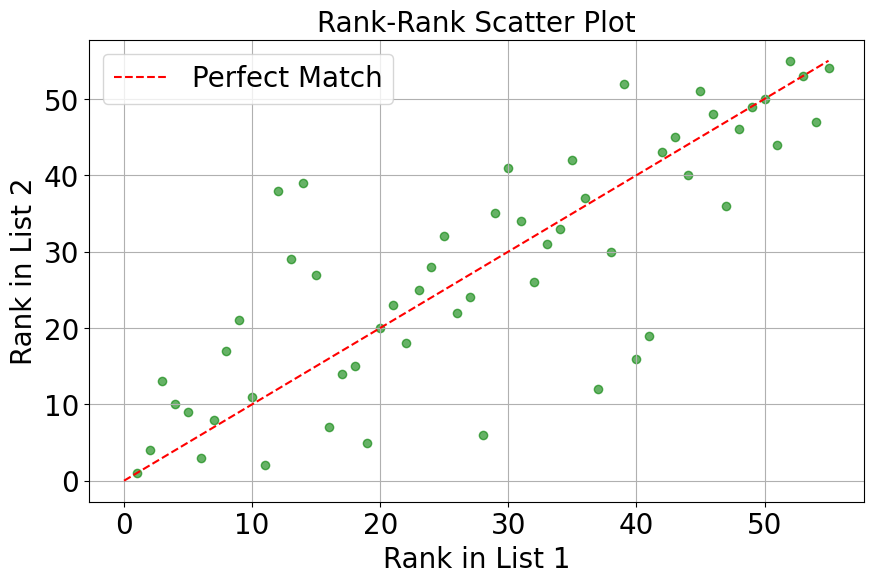}
        \caption{Data Shapley rank scatter-scatter plot with Gaussian process regression and random forest}
    \end{subfigure}
    \hfill
    \begin{subfigure}[b]{0.30\linewidth}
        \centering
        \includegraphics[width=\linewidth]{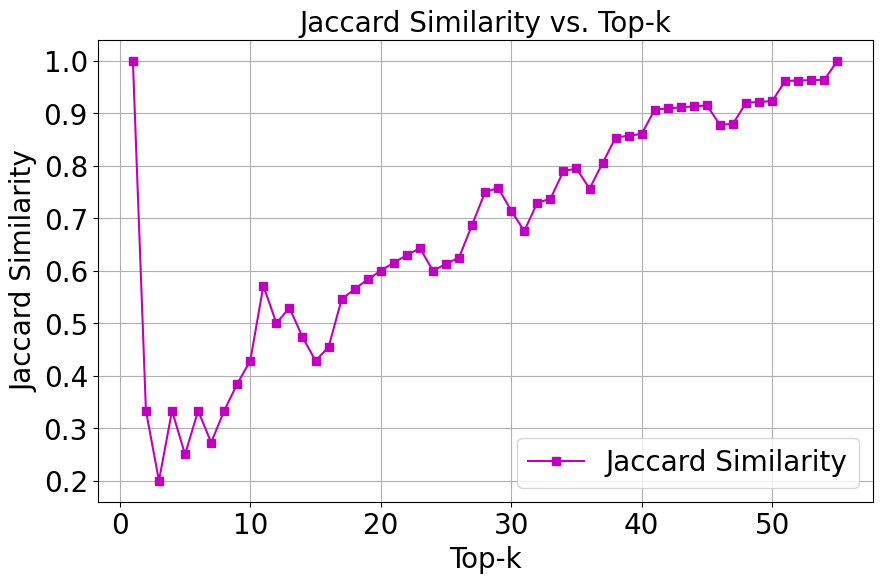}
        \caption{Data Shapley rank Jaccard similarity with Gaussian process regression and random forest}
    \end{subfigure}
    \caption{Effect of model architecture on data valuation framework \edit{($N=87$)}; see Section \ref{ssec:1_gp_rf}.}
    \label{fig:gp_rf}
\end{figure*}
\begin{figure*}[h!]
\centering
    \begin{subfigure}[t]{0.35\linewidth}
        \centering
        \includegraphics[width=0.95\linewidth]{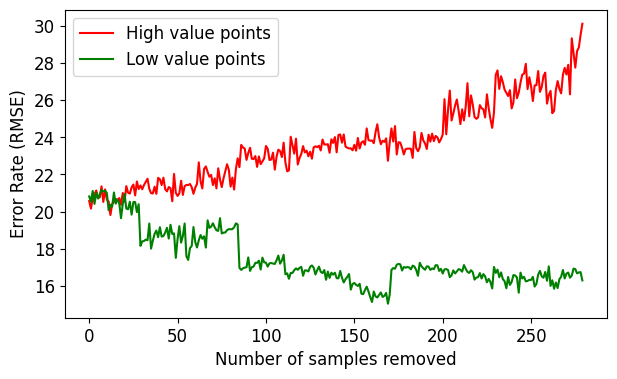}
        \caption{Error plot for median-based imputation}
        \label{fig:error_median}
    \end{subfigure}
    \begin{subfigure}[t]{0.35\linewidth}
        \centering
        \includegraphics[width=\linewidth]{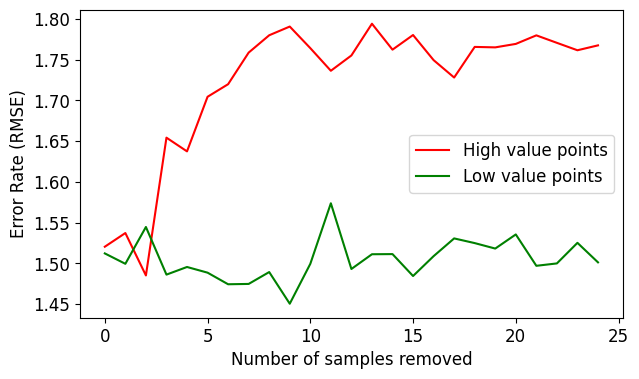}
        \caption{Error plot for listwise deletion}
        \label{fig:error_listwise}
    \end{subfigure}
    \begin{subfigure}[b]{0.35\linewidth}
        \centering
        \includegraphics[width=\linewidth]{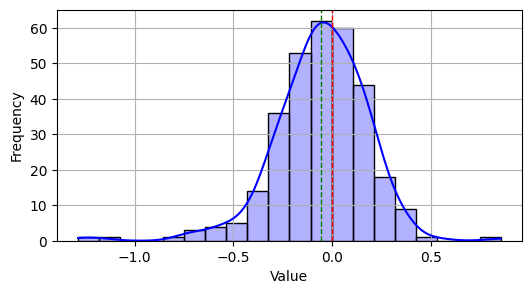}
        \caption{Data Shapley value distribution(median-based imputation)}
        \label{fig:median_dist}
    \end{subfigure}
    \begin{subfigure}[b]{0.35\linewidth}
        \centering
        \includegraphics[width=\linewidth]{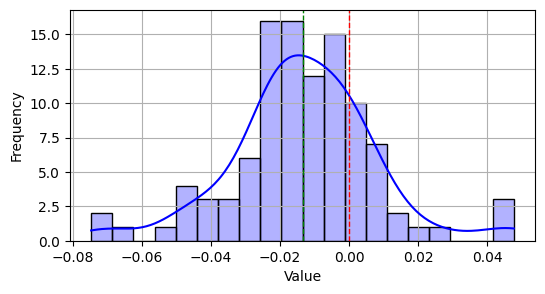}
        \caption{Data Shapley value distribution for listwise deletion}
        \label{fig:listwise_dist}
    \end{subfigure}
    \caption{Enhancing missing data strategies through data valuation \edit{($N=491$)}; see Section \ref{ssec:2_imputation}.}
    \label{fig:imputation}
\end{figure*}
\begin{figure*}[t]
\centering
    \begin{subfigure}[t]{0.30\linewidth}
        \centering
        \includegraphics[width=\linewidth]{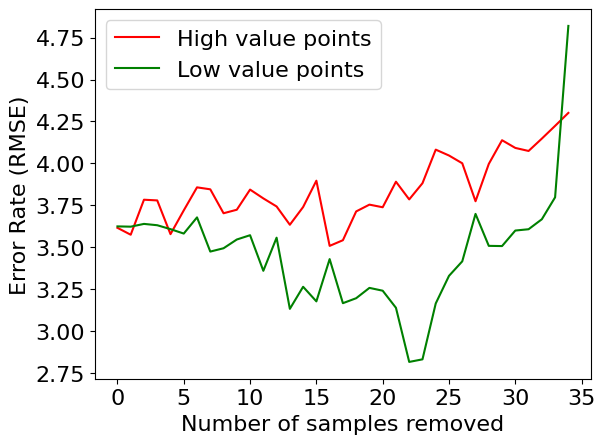}
        \caption{Error plot for backward direction}
        \label{fig:usa_rf_reverse}
    \end{subfigure}
    \hfill
    \begin{subfigure}[t]{0.30\linewidth}
        \centering
        \includegraphics[width=\linewidth]{figures/usa_gp/rf_error.png}
        \caption{Error plot for random forest on forward direction}
        \label{fig:usa_rf_fwd}
    \end{subfigure}
    \hfill
    \begin{subfigure}[t]{0.30\linewidth}
        \centering
        \raisebox{0.20\height}{\includegraphics[width=\linewidth]{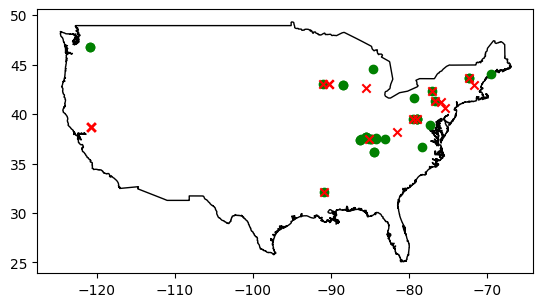}}
        \caption{Data selection on USA map with backward direction (perturbed for privacy)}
        \label{fig:usa_map_rf}
    \end{subfigure}
    \begin{subfigure}[b]{0.30\linewidth}
        \centering
        \includegraphics[width=\linewidth]{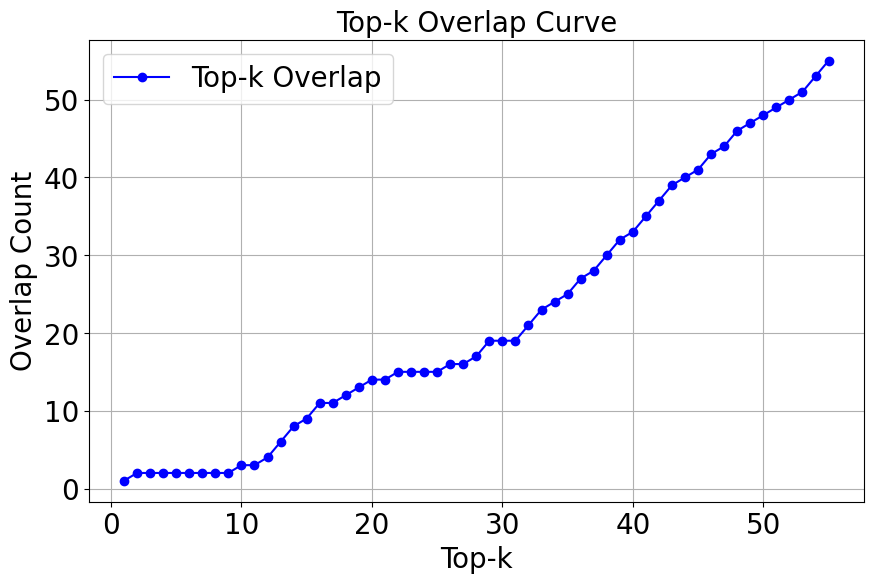}
        \caption{Data Shapley rank overlap for forward vs backward direction}
    \end{subfigure}
    \hfill
    \begin{subfigure}[b]{0.30\linewidth}
        \centering
        \includegraphics[width=\linewidth]{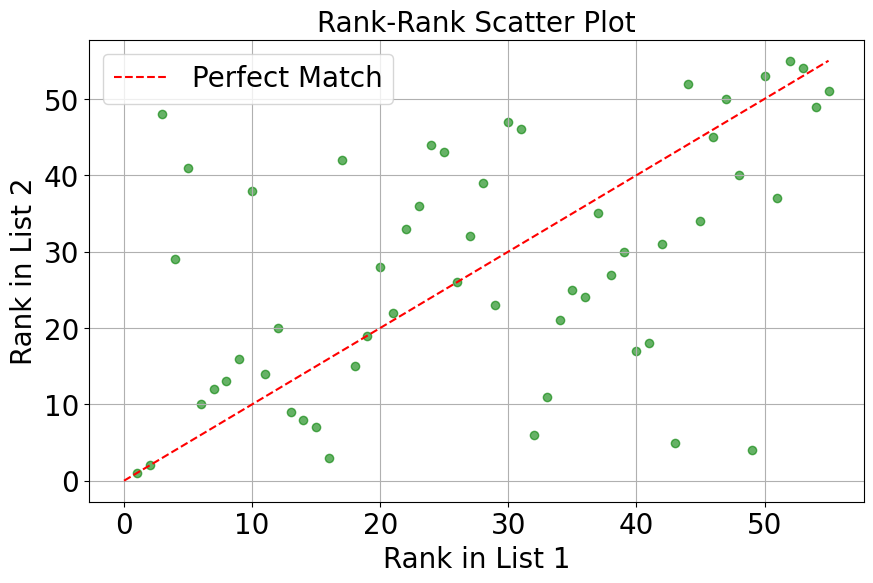}
        \caption{Data Shapley rank scatter-scatter plot for forward vs backward direction}
    \end{subfigure}
    \hfill
    \begin{subfigure}[b]{0.30\linewidth}
        \centering
        \includegraphics[width=\linewidth]{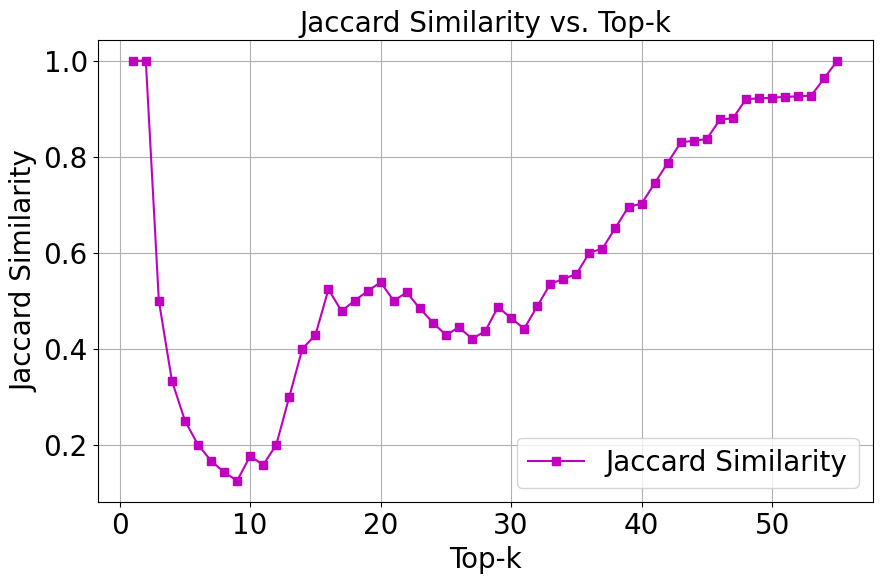}
        \caption{Data Shapley rank Jaccard similarity for forward vs backward direction}
    \end{subfigure}
    \caption{\edit{Random Forest-based data valuation on USA only data ($N=87$); see Section \ref{ssec:3_direction}.}}
    \label{fig:direction_usa}
\end{figure*}
\begin{figure*}[h!]
\centering
    \begin{subfigure}[t]{0.30\linewidth}
        \centering
        \includegraphics[width=\linewidth]{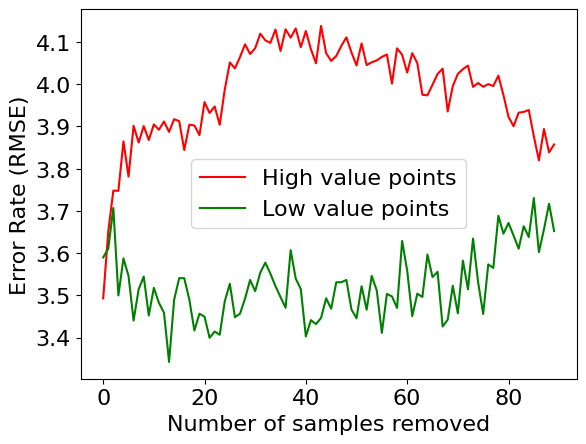}
        \caption{Error plot for backward direction}
    \end{subfigure}
    \hfill
    \begin{subfigure}[t]{0.30\linewidth}
        \centering
        \includegraphics[width=\linewidth]{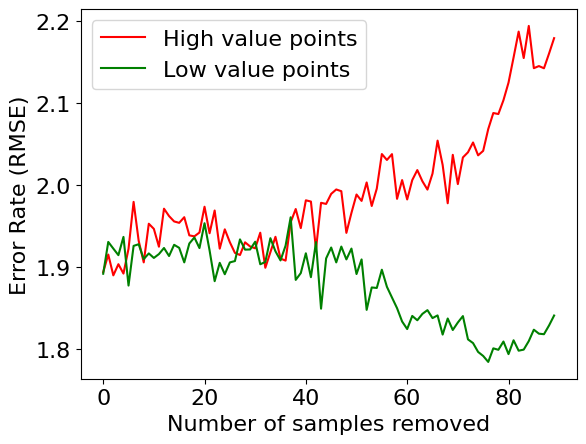}
        \caption{Error plot for random forest on forward direction}
    \end{subfigure}
    \hfill
    \begin{subfigure}[t]{0.25\linewidth}
        \centering
        \includegraphics[trim={2cm 2cm .5cm 4cm},clip, width=\linewidth]{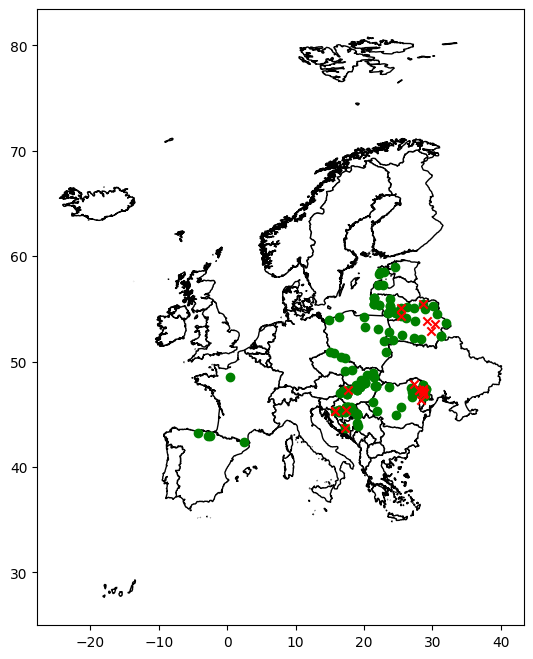}
        \caption{Data selection on the map of Europe with backward direction (perturbed for privacy)}
    \end{subfigure}
    \begin{subfigure}[b]{0.30\linewidth}
        \centering
        \includegraphics[width=\linewidth]{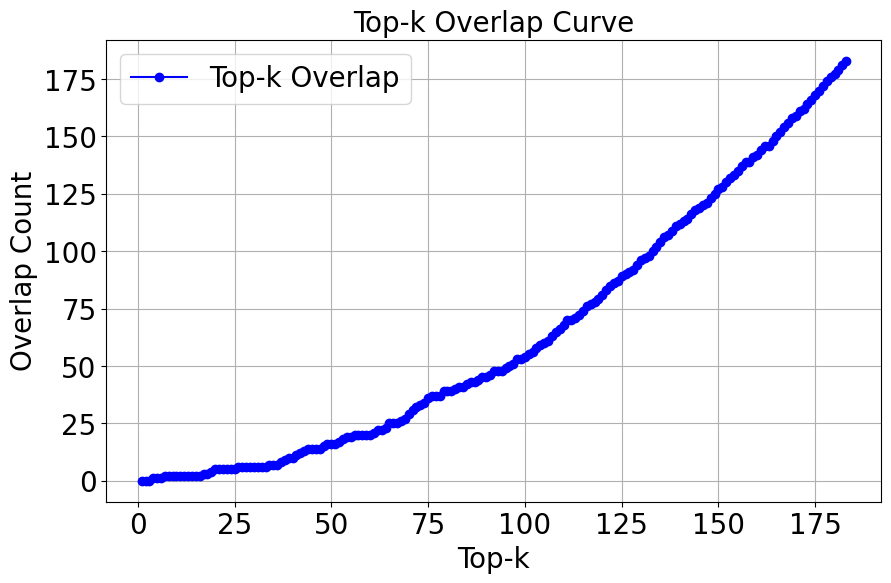}
        \caption{Data Shapley rank overlap for forward vs backward direction}
    \end{subfigure}
    \hfill
    \begin{subfigure}[b]{0.30\linewidth}
        \centering
        \includegraphics[width=\linewidth]{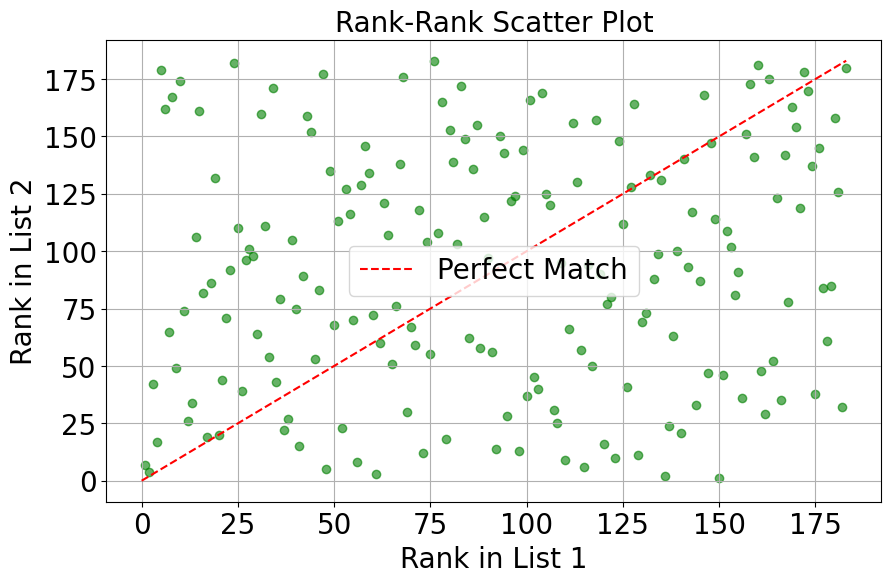}
        \caption{Data Shapley rank scatter-scatter plot for forward vs backward direction}
    \end{subfigure}
    \hfill
    \begin{subfigure}[b]{0.30\linewidth}
        \centering
        \includegraphics[width=\linewidth]{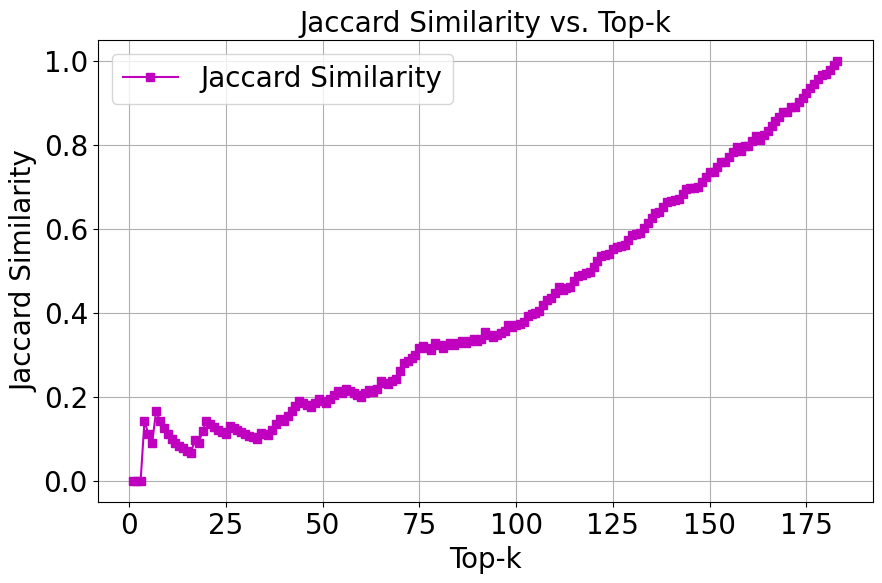}
        \caption{Data Shapley rank Jaccard similarity for forward vs backward direction}
    \end{subfigure}
    \caption{\edit{Random Forest-based data valuation on Europe only data ($N=287$); see Section \ref{ssec:3_direction}.}}
    \label{fig:direction_eu}
\end{figure*}
\begin{figure*}[t]
\centering
    \begin{subfigure}[t]{0.30\linewidth}
        \centering
        \includegraphics[width=\linewidth]{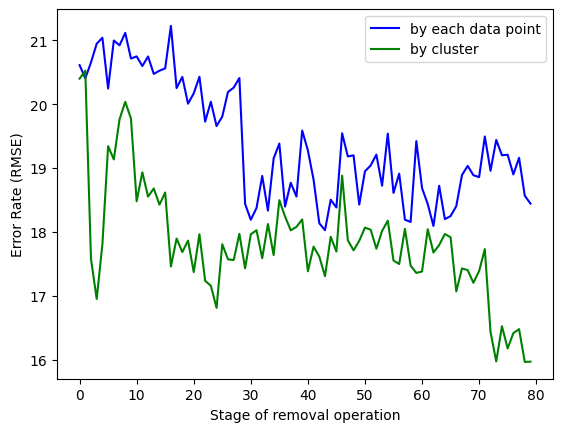}
        \caption{Error plot for removing low-value data points}
        \label{fig:location_unimp}
    \end{subfigure}
    \hfill
    \begin{subfigure}[t]{0.30\linewidth}
        \centering
        \includegraphics[width=\linewidth]{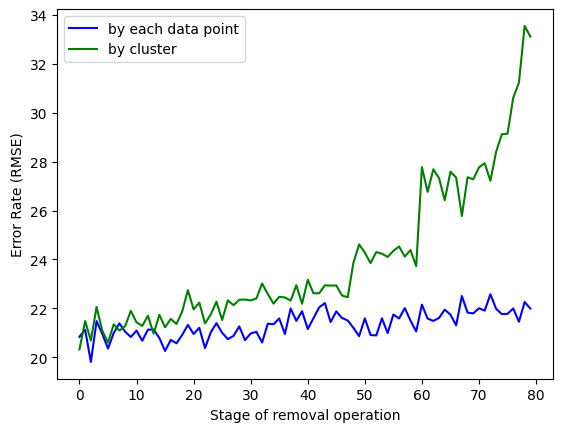}
        \caption{Error plot for removing high-value data points}
        \label{fig:location_imp}
    \end{subfigure}
    \hfill
    \begin{subfigure}[t]{0.30\linewidth}
        \centering
            \includegraphics[width=\linewidth]{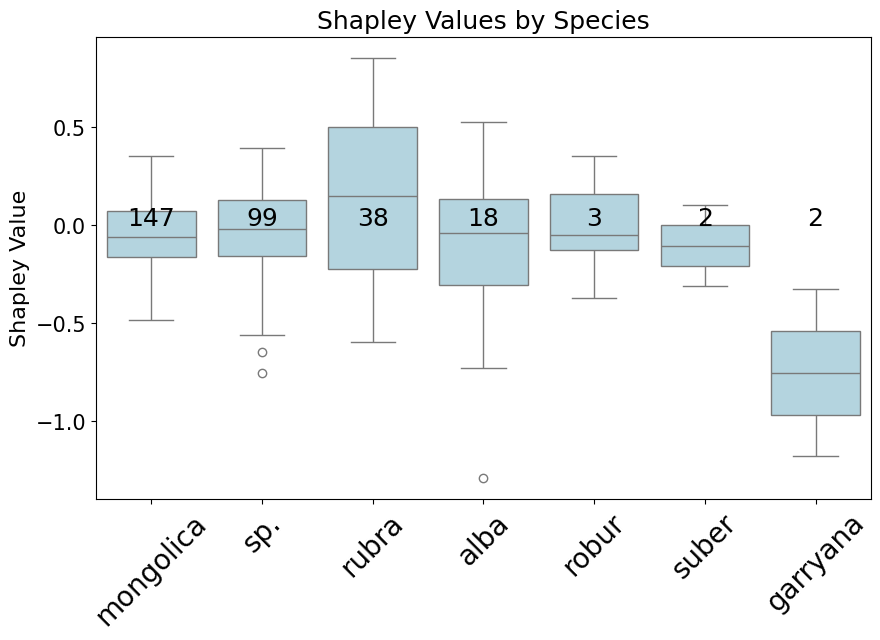}
          \caption{Data Shapley value distribution per species (Section \ref{ssec:5_species})}
          \label{fig:species_dist}
    \end{subfigure}
    \caption{\edit{Optimal Granularity in data selection \edit{($N=491$)}; see Section \ref{ssec:4_clustering} and Data Shapley value distribution per species (see Section \ref{ssec:5_species}).}}
    \label{fig:clustering}
\end{figure*}

\subsection{RQ1: Role of Data Shapley Values in SIRA}
\label{ssec:0_dist}
For this experiment, we explore the relevance and application of data Shapley values in the context of SIRA. 
Fig.~\ref{fig:dist} 
presents distribution plots for one subset of the dataset, evaluated in both directions of SIRA analysis, i.e., forward and backward, as described in Section \ref{sec:methods}. This analysis highlights a significant variation in data Shapley values depending on the dataset and the machine learning model utilized.
These results indicate that subsets with Shapley values exhibiting greater extremities will correspond to larger performance gains following the data selection process.

\subsection{RQ2: Influence of Model Architecture}
\label{ssec:1_gp_rf}
We conducted two types of experiments, removing both high value and low value data points.
We present the error plots showing the effect of removing these two types of data points for Gaussian process regression and random forest models. 
In both architectures, the RMSE increases (indicating performance degradation) when data points with high Shapley values are removed. 
Conversely, the RMSE decreases (indicating performance improvement) or remains stable when data points with low Shapley values are removed (Figure \ref{fig:gp_error} and Figure \ref{fig:rf_error_gp}). 
Figure \ref{fig:map_gp} illustrates one such case, where performance improves by removing specific data points (marked as red x on the map).
Beyond accuracy, we should point out that
random forests, while lacking the theoretically principled uncertainty quantifications of GPR, are much more scalable.
We further investigate whether the Shapley values from the two architectures agree with each other. 
Figures \ref{fig:gp_rf}(d-f) present three types of rank comparison plots, all of which demonstrate a high degree of agreement between the two models. 
This result suggests that data value ranks remain consistent across architectures, reinforcing the robustness of our valuation framework even when computational constraints necessitate different modeling choices.

\subsection{RQ3: How does Shapley-based data selection compare against naive or exhaustive baselines?}
\label{ssec:RQ3_baseline}
To contextualize the effectiveness of our Shapley-based data valuation framework, we compare it against two natural baselines—random removal of data points and leave-one-out (LOO) removal. Using USA-only data (which already captures the main performance trends observed across our study), Table~\ref{tab:baseline} shows that our Truncated Monte Carlo (TMC) Shapley approach delivers the largest improvement in predictive accuracy, reducing RMSE by 0.8459 compared to the initial model, nearly four times greater than the improvement achieved with random removal and significantly outperforming LOO.

\begin{table*}[h]
\centering
\begin{tabular}{lcccc}
\hline
\textbf{Method} & \textbf{Initial RMSE} & \textbf{Best RMSE} & \textbf{\edit{Points Removed}} & $\Delta$ \textbf{RMSE} \\
\hline
Random Removal & 3.6101 & 3.4208 & 2 & 0.1893 \\
Remove Low-Value with LOO & 3.6101 & 3.2435 & 24 & 0.3665 \\
Remove Low-Value with TMC-Shapley (ours) & 3.6101 & \textbf{2.8076} & 22 & \textbf{0.8025} \\
\hline
\end{tabular}
\caption{Comparison of data selection strategies. Shapley-based removal consistently outperforms random removal and achieves greater RMSE reduction than LOO, while being more theoretically grounded and computationally efficient. (Section \ref{ssec:RQ3_baseline})}
\label{tab:baseline}
\end{table*}

\subsection{RQ4: Data Valuation to support Missing Data Imputation}
\label{ssec:2_imputation}
A critical but often underappreciated issue is that imputation strategies implicitly encode assumptions about isotopic stationarity, e.g., median imputation presumes that the conditional distribution of missing values is homogeneous across all spatial and taxonomic contexts (an assumption that may be violated under latitudinal gradients or localized environmental variation). Similarly, listwise deletion assumes that missingness is independent of the isotopic signal (which is rarely true in practice, as remote sites are both more difficult to sample and may exhibit distinct isotopic patterns). These hidden assumptions complicate the interpretation of downstream model performance and underscore the need for principled data valuation frameworks to mitigate imputation-induced biases. To explore this, we conduct experiments with two common approaches for handling missing data: (i) median imputation, replacing missing values with the dataset median, and (ii) listwise deletion, excluding data points containing missing values from training; in addition, we perform data selection experiments by removing both high-value and low-value data points and examining the resulting performance rankings.

We conducted experiments involving the removal of both high-value and low-value data points. 
The error plots (Figures \ref{fig:error_median} and \ref{fig:error_listwise}) illustrate the effects of removing these data points under both missing data handling strategies. For both strategies, the RMSE increases (indicating performance degradation) when data points with high Shapley values are removed. Conversely, the RMSE decreases (indicating performance improvement) or remains stable when data points with low Shapley values are removed.
We observe significant performance improvements when low-value data points are removed from both strategies, with the improvement being more pronounced for median imputation (26.66\%) compared to listwise deletion (5.88\%). This result demonstrates the effectiveness of data valuation as a strategy for improving missing data handling. Moreover, the Shapley values for the median imputation strategy exhibit higher magnitudes compared to listwise deletion (Figures \ref{fig:median_dist} and \ref{fig:listwise_dist}), which aligns with the greater performance improvements observed for median imputation.

\subsection{RQ5: Data Selection Methods for Forward and Backward directions in SIRA}
\label{ssec:3_direction}
In this experiment, we investigate whether selecting data based on valuation metrics can improve performance in both directions of SIRA. For this analysis, we present the results of applying the random forest model in both forward and backward directions using two distinct datasets: USA-only data and Europe-only data.

\textbf{USA-only data:} We present the error plots showing the effect of removing high-value versus low-value data points on the performance of the random forest model. In both directions, the RMSE increases (indicating performance degradation) when data points with high Shapley values, are removed. Similarly, the RMSE either decreases (indicating performance improvement) or remains stable when data points with low Shapley values, are removed (Figures \ref{fig:usa_rf_reverse} and \ref{fig:usa_rf_fwd}). Figure \ref{fig:usa_map_rf} illustrates a specific case where performance improves following the removal of certain data points, marked as red x on the map.

We observe performance improvements when low-value data points are removed from the dataset, with the improvement being more pronounced in the backward direction (27.13\%) compared to the forward direction (0.14\%). This finding demonstrates that performance improvement can indeed vary depending on the direction of SIRA. Moreover, consistent with the observations made in Section \ref{ssec:2_imputation}, the Shapley values associated with the backward direction exhibit higher magnitudes compared to the forward direction (Figure \ref{fig:dist}), which aligns with the greater performance improvements observed for the backward direction.

We further investigate whether the Shapley values from both directions show agreement. To this end, we present three types of rank comparison plots in Figure \ref{fig:direction_usa}(d-f). All three representations indicate a high degree of agreement between the two directions, suggesting that data value ranks remain relatively consistent across different directions of SIRA. Similar results for the Europe-only
dataset are shown in
Fig.~\ref{fig:direction_eu}(a-f).
\subsection{RQ6: Optimal Granularity in Data Selection}
\label{ssec:4_clustering}
Here, we analyze the appropriate level of granularity for data selection to maximize performance when using a specific model and a data valuation framework. Instead of removing one data point at a time, we consider clusters of fixed distances (in kilometers) and remove all data points within that distance if a data point is selected for removal during the data selection process. We performed this location-based data selection for both high value (Figure \ref{fig:location_imp}) and low value (Figure \ref{fig:location_unimp}) data points to understand the impact on model performance.
Consistent with the results of previous experiments, we observe performance improvements when low-value data points are removed. Furthermore, the cluster-based removal approach enhances the model's performance more effectively than the one-by-one removal approach.
\subsection{RQ7: Species-Specific Data Shapley Values and Their Implications}
\label{ssec:5_species}
For this experiment, we explore the variations in data Shapley values across different genera and species, aiming to identify the most and least significant contributors within the dataset. We present a sample distribution plot of data Shapley values for individual species within one dataset (Figure \ref{fig:species_dist}). The plot demonstrates that certain species exhibit significantly higher data Shapley values compared to others, indicating greater contribution to the model's performance.

\section{Deployment Details}
WFID has operationalized the SIRA approach
in this paper to trace the geographic origin of forest products by guiding the collection of high-risk reference samples \cite{birch_wf_2}, analyzing them in certified laboratories \cite{fera_science_3}, and deploying trained geographic origin models via the World Forest ID Evaluation Platform \cite{wf_evaluation_4}, which is used by regulators, certification bodies, auditors, and companies to verify sourcing claims across timber and other EU Deforestation Regulation-relevant commodities such as soy, cacao \cite{wf_bean_5}, and coffee. In a recent timber market study \cite{guardian_6, wf_timber_market_7}, corporate partners used the WFID platform to assess sourcing claims for 59 wood products, focusing on birch due to concerns over sourcing from sanctioned regions, where SIRA and spatial models were used to validate or refute claimed harvest locations. WFID has also supported enforcement efforts, helping identify over 260 tons of allegedly illegal timber \cite{ft_8} in Belgium and supporting at least nine additional ongoing investigations. 
\section{Conclusion and Future Work}
Our data valuation methods demonstrate promising results in SIRA analytics, and region-based data selection further enhances performance by removing low-value data clusters. The greatest gains are observed when low-value data points with higher absolute Shapley values are removed, while the largest drops occur when high-value points are excluded, showing that the inferred values are meaningful for product provenance verification.
\edit{Our analyses further indicate that Shapley-based valuation captures broader spatial and distributional informativeness across regions and species, rather than merely filtering noisy samples, reinforcing its robustness and interpretability for provenance modeling.}
\edit{Future work will generalize these methods across natural resource supply chains, incorporating cross-modal data and exploring greater model scalability and robustness.}

\section*{Acknowledgments}
This paper is based upon work supported by the NSF under Grant No. CMMI-2240402. JT is supported by the Swedish Foundation for Strategic Environmental Research MISTRA (Utmana program). Ruoxi Jia and the ReDS lab also acknowledge support from the National Science Foundation through grants IIS-2312794, IIS-2313130, and OAC-2239622. Any opinions, findings, and conclusions or recommendations expressed in this material are those of the author(s) and do not necessarily reflect the views of the sponsors.
\bibliography{aaai2026}
\appendix
\section{Technical Appendix}
\begin{algorithm}
    \caption{Truncated Monte Carlo Shapley \cite{ghorbani2019data}.}
    \label{alg:tmc shapley}
    \begin{algorithmic}
    \Require datasets $D$ and $T$, function $h$, performance metrics $v$
    \Ensure Shapley values $\phi_i, \forall i$ 
    
       \noindent $\text {Initialize } \phi_i=0, \forall i \text { and } t=0$
       \While{ not converged  }\\
            $t \leftarrow t+1$\\
            $\pi^t \leftarrow \text { Random permutation of train data points }$\\
            $v_0^t \leftarrow v(h, \emptyset, T)$
            \ForAll  {$j=1$ to $N$}
                \If {$\text{ if }\left|v(D)-v_{j-1}^t\right|<\text { Performance Tolerance}$}
                \State  $v_j^t=v_{j-1}^t$ 
                \Comment{Reuse Previous Subset Performance}
                \Else
                \State  $v_j^t= v\left(h, P^{\pi^t}_{j} \cup \pi^t[j], T\right)$
                \Comment{Subset Performance}
                \EndIf
                \State $\phi_{\pi^t[j]} \leftarrow \frac{t-1}{t} \phi_{\pi^{t}[j]}+\frac{1}{t}\left(v_j^t-v_{j-1}^t\right)$ \Comment{Update Value}
            \EndFor
       \EndWhile\\
       \Return{$\phi_i, \forall i$}
    \end{algorithmic}
\end{algorithm}

\begin{algorithm}
        \caption{Data Selection with Iterative Shapley.}
    \label{alg:data select}
    \begin{algorithmic}
    \Require full datasets $D$ and $T$, function $h$, performance metrics $v$
    \Ensure selected training dataset $D'$
    
       \noindent $\text {Initialize } D' \leftarrow D$
       \State $\phi_i \leftarrow \text{Compute Data Values for }D'$ \Comment{Algorithm~\ref{alg:tmc shapley}}
       \While{ $v(h, D', T)$ not converged  } 
            \State $\text{min\_i} \leftarrow \arg \min_i \phi_i$ \Comment{Find $z_i$ with Lowest Value}
            \State $D' \leftarrow D' \setminus \{z_{\text{min\_i}}\}$ \Comment{Remove $z_i$ with Lowest Value}
       \EndWhile\\
       \Return{$D'$}
    \end{algorithmic}
\end{algorithm}

\begin{algorithm}[t]
    \caption{Efficient Monte Carlo Beta Shapley.}
    \label{alg:beta_shapley}
    \begin{algorithmic}[1]
    \Require Datasets $D = \{z_1, \dots, z_n\}$, $T$, function $h$, performance metrics $v$, Beta parameters $\alpha, \beta$.
    \Ensure Beta Shapley values $\psi_i, \forall i \in [n]$.
    
    \State Initialize values $\psi_i = 0, \forall i \in [n]$.
    \State Initialize iteration counter $B=1$.
    
    \ForAll{$k=1$ to $n$} \Comment{Pre-compute normalized weights}
        \State $\tilde{w}_k \leftarrow n \binom{n-1}{k-1}^{-1} \frac{\text{Beta}(k+\beta-1, n-k+\alpha)}{\text{Beta}(\alpha, \beta)}$
    \EndFor

    \While{not converged}
        \ForAll{$j=1$ to $n$}
            \State Sample a cardinality $k$ uniformly from $\{1, \dots, n\}$.
            \State Sample a subset $S \subset D \setminus \{z_j\}$ with $|S| = k-1$ uniformly at random.
            \State $\Delta \leftarrow v(h, S \cup \{z_j\}, T) - v(h, S, T)$. \Comment{Calculate marginal contribution}
            \State $\psi_j \leftarrow \frac{B-1}{B} \psi_j + \frac{1}{B} \cdot \tilde{w}_k \cdot \Delta$. \Comment{Update value with Beta weight}
        \EndFor
        \State $B \leftarrow B+1$.
    \EndWhile \\
    \Return{$\psi_i, \forall i \in [n]$}
    \end{algorithmic}
\end{algorithm}
\end{document}